\documentclass{article}
\usepackage{booktabs}
\usepackage{tablefootnote}
\usepackage{amsmath}
\usepackage{graphicx}
\usepackage[square,sort,comma,numbers]{natbib}

\makeatletter

\providecommand{\tabularnewline}{\\}

\usepackage[preprint]{nips_2018}

\usepackage[utf8]{inputenc} 
\usepackage[T1]{fontenc}    
\usepackage{hyperref}       
\usepackage{url}            
\usepackage{booktabs}       
\usepackage{amsfonts}       
\usepackage{nicefrac}       
\usepackage{microtype}      
\usepackage{svg}

\title{Gated Graph Recursive Neural Networks for Molecular Property Prediction}

\author{%
  Hiroyuki Shindo \\
  Division of Information Science \\
  Nara Institute of Science and Technology \\
  \texttt{shindo@is.naist.jp}
  \And
  Yuji Matsumoto \\
  Division of Information Science \\
  Nara Institute of Science and Technology \\
  \texttt{yuji@is.naist.jp}
}

\@ifundefined{showcaptionsetup}{}{%
 \PassOptionsToPackage{caption=false}{subfig}}
\usepackage{subfig}
\makeatother

\begin{document}
\maketitle
\begin{abstract}
Molecule property prediction is a fundamental problem for computer-aided drug discovery and materials science. Quantum-chemical simulations such as density functional theory (DFT) have been widely used for calculating the molecule properties, however, because of the heavy computational cost, it is difficult to search a huge number of potential chemical compounds. Machine learning methods for molecular modeling are attractive alternatives, however, the development of expressive, accurate, and scalable graph neural networks for learning molecular representations is still challenging. In this work, we propose a simple and powerful graph neural networks for molecular property prediction. We model a molecular as a directed complete graph in which each atom has a spatial position, and introduce a recursive neural network with simple gating function. We also feed input embeddings for every layers as skip connections to accelerate the training. Experimental results show that our model achieves the state-of-the-art performance on the standard benchmark dataset for molecular property prediction.
\end{abstract}

\section{Introduction}

Predicting the properties of molecules is a crucial ingredient for
computer-aided drug discovery and materials development with desired
properties~\cite{kearnes_molecular_2016,gilmer_neural_2017,chmiela_machine_2017,schutt_schnet:_2017}.
Currently, quantum-chemical simulations based on density functional
theory (DFT)~\cite{becke_densityfunctional_1993,burke_perspective_2012}
are widely used to calculate the electronic structure and properties
of molecules. However, because of the heavy computational cost of
DFT, they are difficult to extensively explore the huge number of
potential chemical compounds~\cite{gilmer_neural_2017}. To enlarge
the search space, much effort has been made to apply machine learning
techniques for learning molecular representations in cheminformatics
and materials informatics, while it has not been fully developed.
Efficient and accurate machine learning methods for the prediction
of molecular properties can have a huge impact on the discovery of
novel drugs and materials.

Previous work on molecular modeling with machine learning methods
has mainly focused on developing hand-crafted features for molecular
representations that can reflect structural similarities and biological
activities of molecules. Examples include Extended-Connectivity Fingerprints~\cite{rogers_extended-connectivity_2010},
Coulomb Matrix~\cite{rupp_fast_2012}, Symmetry Function~\cite{behler_generalized_2007},
and Bag-of-Bonds~\cite{hansen_machine_2015}. These molecular representations
can be used for the prediction of molecular properties with logistic
regression and kernel methods.

Recently, deep learning methods have gained a lot of attention for
learning molecular representations, with the availability of large
scale training data generated by quantum-chemical simulations~\cite{gilmer_neural_2017}.
In particular, graph neural networks are reasonable and attractive
approaches since they can learn appropriate molecular representations
that are invariant to graph isomorphism in an end-to-end fashion.
While a number of graph neural networks have been proposed and applied
to molecular modeling, developing accurate and scalable neural networks
enough to express a variety of molecules is still a challenging problem.

In this work, we present a simple and powerful graph neural network,
gated graph recursive neural networks (GGRNet), for learning molecular
representations and predicting molecular properties. To construct
an expressive and accurate neural network, we model a molecule as
a complete directed graph in which each atom has a three-dimensional
coordinates, and update hidden vectors of atoms depending on the distances
between them. In our model, the parameters for learning hidden atom
vectors are shared across all layers, and the input embeddings are
fed into every layers as skip connections to accelerate the training.
Our model also allows to incorporate arbitrary features such as the
number of atoms in the molecule, which is helpful for learning better
representations of molecules.

We validate our model on three benchmark datasets for molecular property
prediction: QM7b, QM8, and QM9 and empirically show that our model
achieves superior performance than conventional methods, which highlights
the potential of our model for molecular graph learning.

\section{Related Work}

\paragraph{Molecular fingerprints}

In cheminformatics, hand-crafted features for molecules, referred
to as molecular fingerprints, have been actively developed for encoding
the structure of molecules~\cite{carhart_atom_1985,rogers_extended-connectivity_2010,rupp_fast_2012,behler_generalized_2007,hansen_machine_2015}.
These molecular fingerprints are typically a binary or integer vector
that represents the presence of particular substructures in the molecule,
and can be used as feature vectors for the prediction of molecular
properties with machine learning~\cite{myint_molecular_2012,elton_applying_2018}.

For example, in Extended-Connectivity Circular Fingerprints (ECFP)~\cite{rogers_extended-connectivity_2010},
atoms are initially assigned to integer identifiers, then the identifiers
are iteratively updated with neighboring atoms and collected into
the fingerprint set. Bag-of-bonds descriptor~\cite{hansen_machine_2015},
inspired by ``bag-of-words'' featurization in natural language processing,
collects ``bags'' that correspond to different types of bonds such
as ``C-C'' and ``C-H'', and each bond in the bag is vectorized
as $Z_{i}Z_{j}/\left|\mathbf{R}_{i}-\mathbf{R}_{j}\right|$ where
$Z_{i}$ and $Z_{j}$ are the nuclear charges, while $\mathbf{R}_{i}$
and $\mathbf{R}_{j}$ are the positions of the two atoms in the bond.

Duvenaud et al.~\cite{duvenaud_convolutional_2015} generalized the
computation of conventional fingerprints to be differentiable and
learnable via backpropagation. They showed that the learnable fingerprints
improve the predictive accuracy of molecules compared with the traditional
fingerprints.

\paragraph{Graph neural networks for molecules}

Recently, graph neural networks have attracted a lot of attention
for a wide variety of tasks, including graph link prediction~\cite{zhang_link_2018},
chemistry and biology~\cite{fout_protein_2017,gilmer_neural_2017,schutt_quantum-chemical_2017},
natural language processing~\cite{rahimi_semi-supervised_2018,sorokin_modeling_2018},
and computer vision~\cite{qi_3d_2017,charles_pointnet:_2017}. The
neural network on graphs was early proposed by Gori et al. \cite{gori_new_2005}
and Scarselli et al. \cite{scarselli_graph_2009}, and a large number
of architectures have been proposed until now.

Gilmer et al. \cite{gilmer_neural_2017} proposed neural message passing
networks (MPNNs) framework for learning molecular representations
and showed that many graph neural networks such as Gated Graph Neural
Networks (GG-NN)~\cite{li_gated_2016}, Interaction Networks~\cite{battaglia_interaction_2016},
and Deep Tensor Neural Networks (DTNN)~\cite{schutt_quantum-chemical_2017}
fall under the framework. They tested the MPNN on QM9 dataset for
the prediction of molecular properties and achieved the state-of-the-art
results.

Sch\"{u}tt et al. \cite{schutt_schnet:_2017} introduced SchNet with
the continuous-filter convolutions that map the atom positions in
the molecule to the corresponding filter values. The learned filters
are interacted with atom features to generate more sophisticated atom
representations. The continuous-filter convolutions are incorporated
into graph neural networks for the prediction of molecular energy
and interatomic forces. SchNet is similar to ours in that it assumes
the atoms have spartial information and learns the interactions between
atoms depending on the distances.

Veli\v{c}kovi\'{c} et al. \cite{velickovic_graph_2018} introduced
an attention mechanism to graph neural networks. The graph attention
networks aggregate the hidden representations of vertices in the graph
by weighting over its neighbors, following the self-attention mechanism
that is widely used in many sequential models. They show that the
attention mechanism is helpful for node classification tasks on citation
network and protein-protein interaction.

\section{Gated Graph Recursive Neural Networks (GGRNet)}

\begin{figure}
\center

\subfloat[]{\includegraphics[scale=0.2]{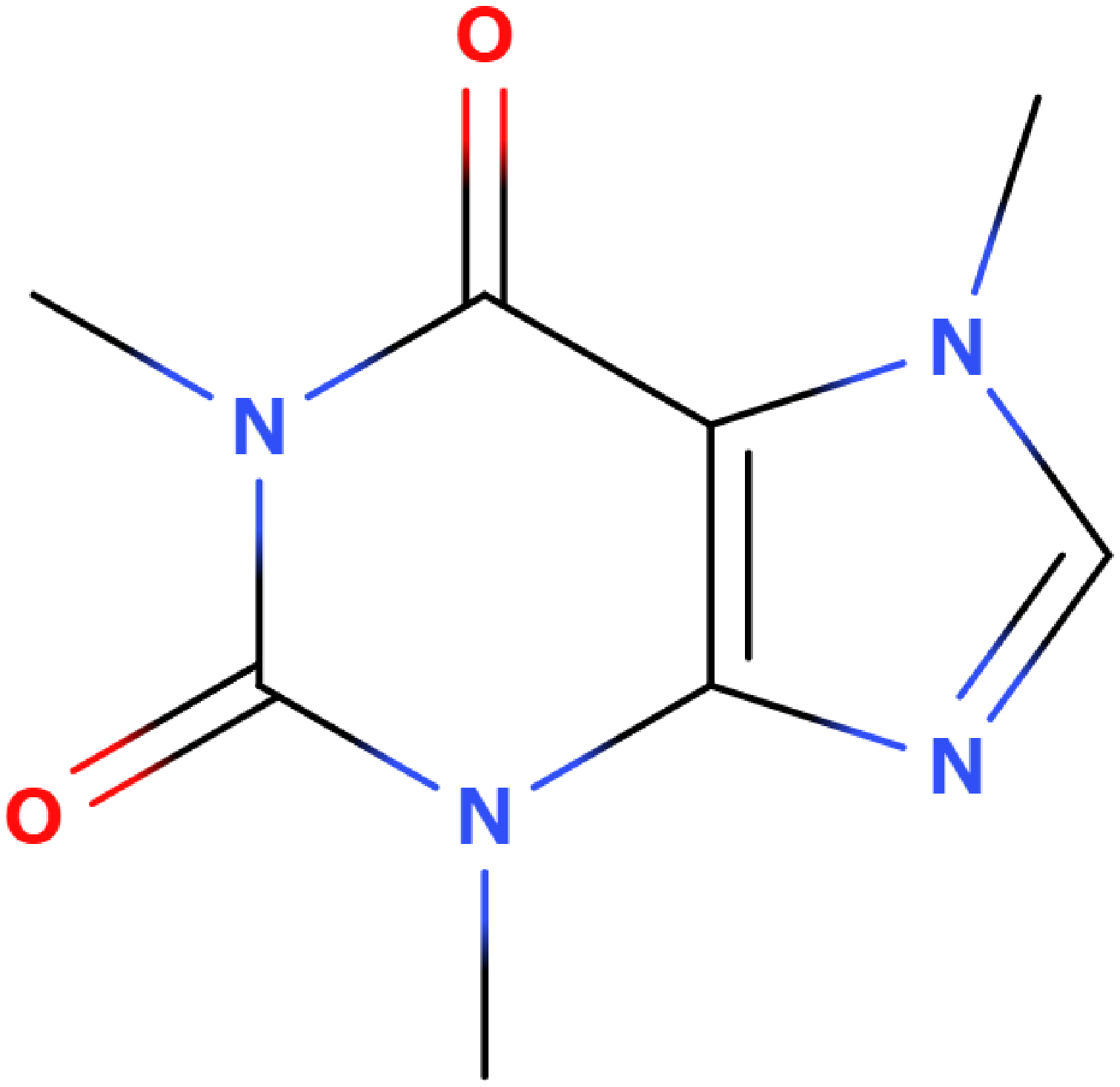}

}~~ ~ ~ ~ ~ ~ ~ ~ ~ ~ ~ ~ \subfloat[]{\includegraphics[scale=0.3]{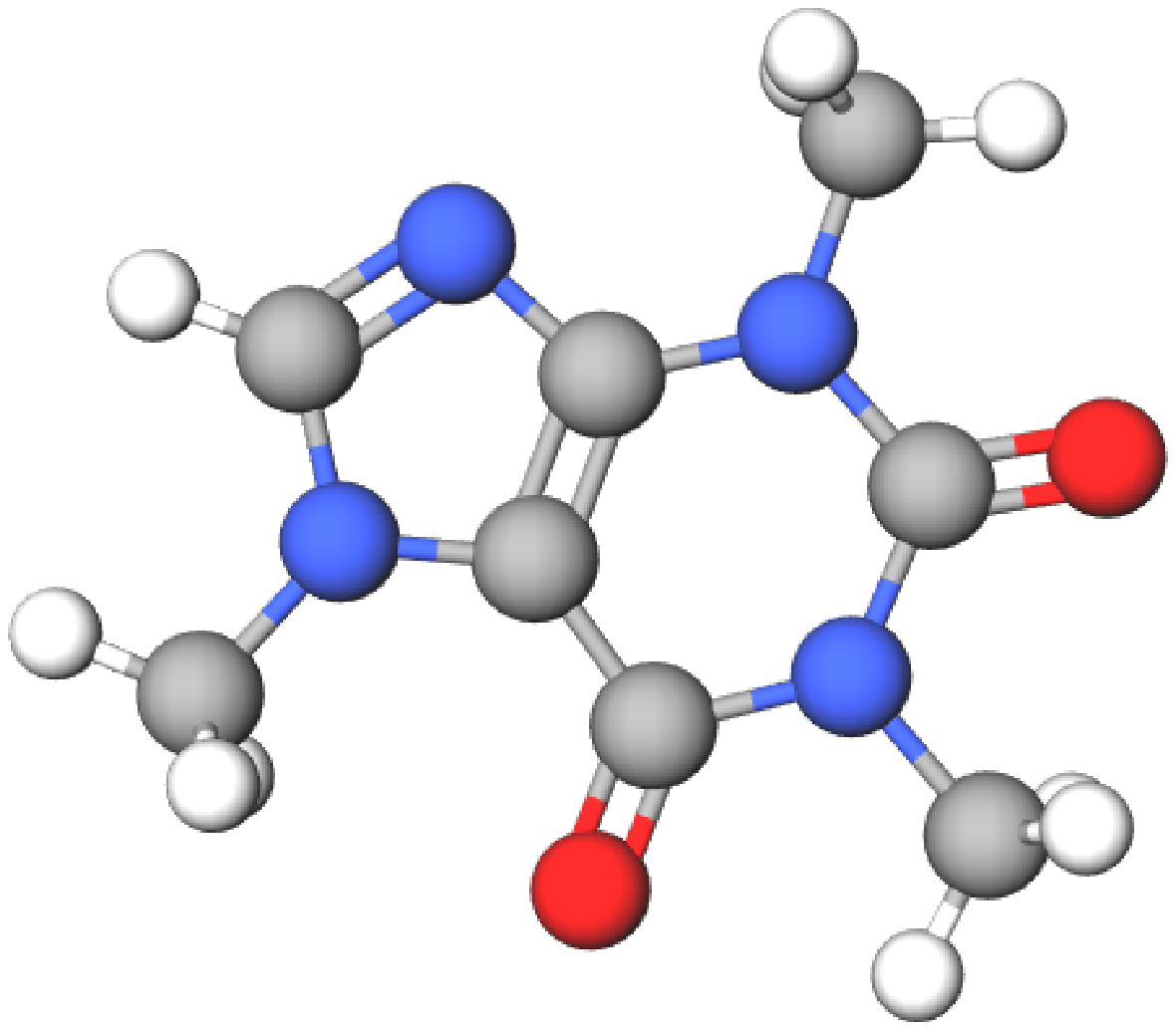}

}

\caption{\label{fig:molecule structures}(a) 2d structure of a molecule. (b)
3d structure of a molecule.}

\end{figure}

A molecular graph consists of atoms and chemical bonds where atoms
correspond to vertices and chemical bonds correspond to edges. Figure
\ref{fig:molecule structures} shows an example of 2D and 3D structures
of a molecule. The conventional graph neural networks mainly handle
with 2D structure while we treat the molecule as 3D graph structure
where three-dimensional coordinates of each atom are given.

\subsection{Model}

\begin{figure}
\center

\includegraphics[scale=0.7]{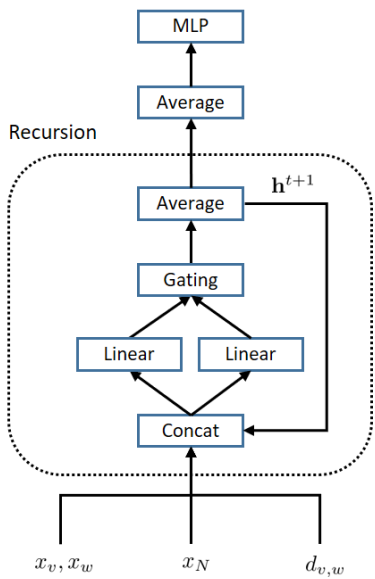}

\caption{\label{fig:GGRNet}Network architecture of GGRNet}
\end{figure}

In this section, we propose a gated graph recursive neural networks
(GGRNet) for accurately learning molecular representations. Since
our model can be formulated as a MPNN framework~\cite{gilmer_neural_2017},
we basically follow the notation and definitions given in \cite{gilmer_neural_2017}.

To build an expressive graph neural network for molecules, one needs
to stack multiple layers for learning hidden representations of atoms
and bonds. However, as the number of parameters increases, it suffers
from inefficient training and lower performance. Our GGRNet alleviates
the problem by the following ideas: 1) the parameters for updating
hidden representations are shared across all layers, 2) the input
representations (atom embeddings and additional input features) are
fed into every layers as skip connections to accelerate the training,
and 3) the feature vector at each atom is updated by using that of
every other atoms in the graph depending on the distances between
atoms.

Formally, suppose we are given a molecular graph $G$ in which each
atom has a three-dimensional position. Let $\mathbf{x}_{v}$ and $\mathbf{x}_{w}$
be d-dimensional atom embeddings of a vertex $v$ and $w$, respectively.
In our model, the hidden vector of vertex $v$ at time $t$, denotes
as$\mathbf{h}_{v}^{t}$, is given by the message function $M$ and
the update function $U$ as follows:

\begin{align*}
\mathbf{m}_{v}^{t+1} & =\sum_{w\neq v}M\left(\mathbf{x}_{v},\mathbf{h}_{v}^{t},\mathbf{x}_{w},\mathbf{h}_{w}^{t}\right)\\
\mathbf{h}_{v}^{t+1} & =U\left(\mathbf{h}_{v}^{t},\mathbf{m}_{v}^{t+1}\right)
\end{align*}

where the values of $M$ are summed over all vertices $w$ except
$v$, which means we assume every pair of vertices in the graph has
an edge and communicate each other for every time-step.

In the original MPNN, the message function $M$ is defined as $M_{t}\left(\mathbf{h}_{v}^{t},\mathbf{h}_{w}^{t}\right)$,
and $w$ as the neighbors of $v$. Our model extends it to always
feed the input vectors$\mathbf{x}_{v}$ and $\mathbf{x}_{w}$ into
the message function. Furthermore, parameters of $M$ and $U$ are
shared across all time-steps.

After the $T$ time-step of message updates, the readout function
$R$ aggregates all the hidden representations:

\[
y=R\left(\left\{ \mathbf{h}_{v}^{T}\left|v\in G\right.\right\} \right)
\]

where $y$ is a target value such as molecular property.

In this work, we consider the following funcion as $M$:

\begin{align*}
\mathbf{p}_{v,w}^{t} & =\mathbf{W}_{p}\cdot\left[\oplus\left(\mathbf{x}_{v},\mathbf{h}_{v}^{t},\mathbf{x}_{w},\mathbf{h}_{w}^{t},\mathbf{x}_{N},\mathbf{d}_{v,w}\right)\right]+\mathbf{b}_{p}\\
\mathbf{q}_{v,w}^{t} & =\mathbf{W}_{q}\cdot\left[\oplus\left(\mathbf{x}_{v},\mathbf{h}_{v}^{t},\mathbf{x}_{w},\mathbf{h}_{w}^{t},\mathbf{x}_{N},\mathbf{d}_{v,w}\right)\right]+\mathbf{b}_{q}\\
\mathbf{m}_{v,w}^{t+1} & =\sigma\left(\mathbf{p}_{v,w}^{t}\right)\odot\tanh\left(\mathbf{q}_{v,w}^{t}\right)\\
\mathbf{m}_{v}^{t+1} & =\sum_{w\neq v}\mathbf{m}_{v,w}^{t+1}
\end{align*}

where $\oplus$ is a concatenation function and $\sigma$ is a sigmoid
function. $\mathbf{W}_{p}$, $\mathbf{W}_{q}$, $\mathbf{b}_{p}$and
$\mathbf{b}_{q}$ are model parameters to be learned, which are shared
across all time-steps. Note that we assume $G$ as a directed graph,
hence $\mathbf{m}_{v,w}^{t+1}\neq\mathbf{m}_{w,v}^{t+1}$.

In addition to $\mathbf{x}_{v}$, $\mathbf{h}_{v}^{t}$, $\mathbf{x}_{w}$,
and $\mathbf{h}_{w}^{t}$, we consider two additional input features:
counting feature $\mathbf{x}_{N}$ and distance feature $\mathbf{d}_{v,w}$.
The counting feature is a real-valued embedding vector that corresponds
to the number of atoms $N$ in the graph. Therefore, the molecules
with the same number of atoms share the same counting embeddings.
The distance feature $\mathbf{d}_{v,w}$ is a one-dimensional vector
whose value $d_{v,w}$ is the reciprocal of the Euclid distance between
$v$ and $w$, that is, $d_{v,w}=1/\sqrt{\left(v_{x}-w_{x}\right)^{2}-\left(v_{y}-w_{y}\right)^{2}-\left(v_{z}-w_{z}\right)^{2}}$
where $v_{x}$, $v_{y}$, $v_{z}$ are three-dimensional coodinates
of $v$ and so as $w$.

Initially, $\mathbf{h}_{v}^{0}=\mathbf{0}$ for all $v$. The gating
function: $\sigma\odot\tanh$, inspired by LSTM and gated convolutional
neural networks~\cite{dauphin_language_2017}, is used to extract
effective features from the previous hidden vectors and input features.

The update function $U$ is simply an average as follows:

\[
\mathbf{h}_{v}^{t+1}=\frac{1}{N}\mathbf{m}_{v}^{t+1}
\]

where $N$ is the number of atoms in the molecule.

Finally, the readout function $R$ is given as follows:

\[
y=\textrm{MLP}\left(\frac{1}{N}\sum_{v\in G}\mathbf{h}_{v}^{T}\right)
\]

where we average all the hidden vectors in the graph, followed by
the standard three-layer MLP with ReLU activation function.

Different from the original MPNN, our model always feeds the same
$\mathbf{x}$ and additional input features into the recursive time
step. The impact of counting feature and distance feature is evaluated
in the experimental section.

The network architecture of GGRNet is shown in figure \ref{fig:GGRNet}.

\section{Experiments}

\begin{table}
\caption{\label{tab:hyperparameters}Hyper-parameter settings of our model
for all experiments.}

\centering
\renewcommand{\arraystretch}{1.3}

\begin{tabular}{cccc}
\toprule 
Hyper-parameters \textbackslash{} Dataset & QM7b & QM8 & QM9\tabularnewline
\midrule 
Atom embedding size & 50 & 50 & 50\tabularnewline
Count embedding size & 50 & 50 & 50\tabularnewline
Hidden vector size & 100 & 100 & 100\tabularnewline
Number of recursive layers $T$ & 5 & 5 & 5\tabularnewline
Learning rate $\alpha_{0}$ & 0.03 & 0.03 & 0.01\tabularnewline
Learning rate decay $k$ & 0.01 & 0.01 & 0.05\tabularnewline
Number of epochs & 500 & 500 & 200\tabularnewline
Batch size & 10 & 10 & 10\tabularnewline
Gradient clipping & 10.0 & 10.0 & 10.0\tabularnewline
\bottomrule
\end{tabular}
\end{table}

We validate the performance of our GGRNet on molecular datasets in
MoleculeNet~\cite{wu_moleculenet:_2018}. MoleculeNet is a comprehensive
benchmark for molecular machine learning and it contains multiple
datasets for regression and classification tasks. In this work, we
use QM7b~\cite{montavon_machine_2013}, QM8~\cite{ramakrishnan_electronic_2015},
and QM9~\cite{ramakrishnan_quantum_2014} datasets for regression
task. In QM8 and QM9, the input is a set of discrete molecular graph
with spatial positions of atoms. In QM7b, only spatial positions of
atoms for each molecule are available. The target is a real-valued
molecule property such as the energy of the electron and the heat
capacity. For details about the dataset, please refer to \cite{wu_moleculenet:_2018}.

We compare our GGRNet against the following three baselines: graph
convolutional models (GC)~\cite{wu_moleculenet:_2018}, deep tensor
neural networks (DTNN)~\cite{schutt_quantum-chemical_2017}, and
MPNN~\cite{gilmer_neural_2017}. MPNN is the implementation of an
edge network as message passing function and a set2set model as readout
function. The edge network considers all neighbor atoms and the feature
vectors of atoms are updated with gated recurrent units. In the readout
phase, LSTM with attention mechanism is applied to a sequence of feature
vectors to generate the final feature vector of the molecule.

These three baseline methods achieve state-of-the-art performance
on a variety of tasks for molecular modeling and outperform the conventional
methods using hand-crafted features. All the baseline results are
taken from MoleculeNet~\cite{wu_moleculenet:_2018}. The codes of
baseline models are publicly available via DeepChem open source library~\cite{ramsundar_deep_2019}.

The hyperparameters of our GGRNet in the experiments are shown in
table \ref{tab:hyperparameters}. As shown in the table, we use almost
the same hyper-parameters for every dataset. Following the MoleculeNet,
we randomly split the dataset into training, validation, and test
as 80/10/10 ratio. All the reported results are averaged over three
independent runs.

Target properties in the training set are normalized to zero mean
and unit variance using only the training set. For evaluation, the
predicted values of the test set are inversely transformed using the
training mean and variance. The loss function is mean squared error
(MSE) between the model output and the target value. For evaluation,
we use mean absolute error (MAE). We use stochastic gradient descent
(SGD) for training our model. The learning rate $\alpha$ is given
by $\alpha=\alpha_{0}/(1+ke)$ where $\alpha_{0}$, $k$, $e$ are
the initial learning rate, the decay rate, and the number of epochs,
respectively. All the parameters including atom embeddings and counting
embeddings are initialized randomly and updated during training.

\subsection{QM7b Dataset}

\begin{table}
\caption{\label{tab:QM7b}QM7b test set performances (MAE)}

\centering
\renewcommand{\arraystretch}{1.3}

\begin{tabular}{cccc}
\toprule 
Property \textbackslash{} Model & Unit & DTNN~\cite{schutt_quantum-chemical_2017} & GGRNet\tabularnewline
\midrule
Atomization energy (PBE0) & kcal / mol & 21.5 & \textbf{13.7}\tabularnewline
Maximal absorption intensity (ZINDO)~\tablefootnote{The MoleculeNet paper incorrectly states this entry as ``Excitation
energy of maximal optimal absorption - ZINDO''} & eV & 1.26 & \textbf{1.02}\tabularnewline
Excitation energy at maximal absorption (ZINDO)~\tablefootnote{The MoleculeNet paper incorrectly states this entry as ``Highest
absorption - ZINDO''} & Arbitrary & 0.074 & \textbf{0.072}\tabularnewline
HOMO (ZINDO) & eV & 0.192 & \textbf{0.140}\tabularnewline
LUMO (ZINDO) & eV & 0.159 & \textbf{0.0915}\tabularnewline
First excitation energy (ZINDO) & eV & 0.296 & \textbf{0.121}\tabularnewline
Ionization potential (ZINDO) & eV & 0.214 & \textbf{0.176}\tabularnewline
Electron affinity (ZINDO) & eV & 0.174 & \textbf{0.0940}\tabularnewline
HOMO (PBE0) & eV & 0.155 & \textbf{0.142}\tabularnewline
LUMO (PBE0) & eV & 0.129 & \textbf{0.092}\tabularnewline
HOMO (GW) & eV & 0.166 & \textbf{0.142}\tabularnewline
LUMO (GW) & eV & 0.139 & \textbf{0.118}\tabularnewline
Polarizability (PBE0) & $\mbox{\AA}^3$ & 0.173 & \textbf{0.100}\tabularnewline
Polarizability (SCS) & $\mbox{\AA}^3$ & 0.149 & \textbf{0.0578}\tabularnewline
\bottomrule
\end{tabular}
\end{table}

QM7b consists of 7,211 small organic molecules with 14 properties,
which is a subset of the GDB-13 database~\cite{blum_970_2009}. Each
molecule consists of Hydrogen (H), Carbon (C), Oxygen (O), Nitrogen
(N), Sulfur (S), and Chlorine (Cl). The three-dimensional coordinates
of the most stable conformation and the electronic properties such
as HOMO, LUMO, and electron affinity for each molecule are provided
calculated by DFT simulation. The discrete graph structure of molecules
are missing. Following \cite{gilmer_neural_2017}, we train our model
per target rather than multi-task learning since the per-target training
gives superior performance than joint training.

Table \ref{tab:QM7b} shows mean absolute error (MAE) on QM7b dataset.
As shown in the table, our GGRNet consistently outperforms DTNN. The
results show that GGRNet is expressive to learn molecular representations
as we expected.

\subsection{QM8 Dataset}

\begin{table}
\caption{\label{tab:QM8}QM8 test set performances (MAE)}

\centering
\renewcommand{\arraystretch}{1.3}

\begin{tabular}{ccccc}
\toprule 
Property \textbackslash{} Model & GC & DTNN & MPNN & GGRNet\tabularnewline
\midrule
E1-CC2 & 0.0074 & 0.0092 & 0.0084 & \textbf{0.0057}\tabularnewline
E2-CC2 & 0.0085 & 0.0092 & 0.0091 & \textbf{0.0058}\tabularnewline
f1-CC2 & 0.0175 & 0.0182 & \textbf{0.0151} & 0.0152\tabularnewline
f2-CC2 & 0.0328 & 0.0377 & \textbf{0.0314} & 0.0347\tabularnewline
E1-PBE0 & 0.0076 & 0.0090 & 0.0083 & \textbf{0.0053}\tabularnewline
E2-PBE0 & 0.0083 & 0.0086 & 0.0086 & \textbf{0.0054}\tabularnewline
f1-PBE0 & 0.0125 & 0.0155 & \textbf{0.0123} & 0.0130\tabularnewline
f2-PBE0 & 0.0246 & 0.0281 & \textbf{0.0236} & 0.0289\tabularnewline
E1-CAM & 0.0070 & 0.0086 & 0.0079 & \textbf{0.0052}\tabularnewline
E2-CAM & 0.0076 & 0.0082 & 0.0082 & \textbf{0.0056}\tabularnewline
f1-CAM & 0.0153 & 0.0180 & \textbf{0.0134} & \textbf{0.0134}\tabularnewline
f2-CAM & 0.0285 & 0.0322 & \textbf{0.0258} & 0.0291\tabularnewline
\bottomrule
\end{tabular}
\end{table}

QM8 dataset~\cite{ramakrishnan_quantum_2014} is also a part of GDB-13
database~\cite{blum_970_2009} and contains 21,786 small organic
molecules with 12 properties. In QM8, the time-dependent density functional
theory (TDDFT) and second-order approximate coupled-cluster (CC2)
are applied to calculate the molecular properties. As with QM7b, we
train our model per target.

The results are shown in table \ref{tab:QM8}. As shown in the table,
our GGRNet achieved the best results on 7/12 cases. However, in some
cases such as ``f1-CC2'' and ``f2-CC2'', our model gets stuck
and suffers from inefficient training. We believe increasing the number
of epochs slightly improves the performance, however, more expressive
neural architecture is required to learn better molecular representations
with a small number of epochs. We hypothesize that tbe gating function
of GGRNet may be the cause of gradient vanishing. One solution is
to add batch normalization or other normalization layers to GGRNet.

In ``f1'' and ``f2'' dataset, MPNN achieves the best among all
other methods. We do not have a clear explanation of the results at
present, however, there might be an essential differences between
MPNN and GGRNet. One essential difference between MPNN and GGRNet
is that MPNN employs expressive readout function by using LSTM, while
GGRNet uses simple average function.

\subsection{QM9 Dataset}

\begin{table}
\caption{\label{tab:QM9}QM9 test set performances (MAE)}

\centering
\renewcommand{\arraystretch}{1.3}

\begin{tabular}{cccccc}
\toprule 
Property \textbackslash{} Model & Unit & GC & DTNN & MPNN & GGRNet\tabularnewline
\midrule
mu & Debye & 0.583 & 0.244 & 0.358 & \textbf{0.172}\tabularnewline
alpha & $\textrm{Bohr}^{3}$ & 1.37 & 0.95 & 0.89 & \textbf{0.453}\tabularnewline
HOMO & Hartree & 0.00716 & 0.00388 & 0.00541 & \textbf{0.00372}\tabularnewline
LUMO & Hartree & 0.00921 & 0.00513 & 0.00623 & \textbf{0.00410}\tabularnewline
gap & Hartree & 0.0112 & 0.0066 & 0.0082 & \textbf{0.00536}\tabularnewline
R2 & $\textrm{Bohr}^{2}$ & 35.9 & 17.0 & 28.5 & \textbf{1.61}\tabularnewline
ZPVE & Hartree & 0.00299 & 0.00172 & 0.00216 & \textbf{0.000861}\tabularnewline
U0 & kcal/mol & 3.41 & 2.43 & 2.05 & 31.1\tabularnewline
U & kcal/mol & 3.41 & 2.43 & 2.00 & 28.0\tabularnewline
H & kcal/mol & 3.41 & 2.43 & 2.02 & 34.5\tabularnewline
G & kcal/mol & 3.41 & 2.43 & 2.02 & 30.6\tabularnewline
Cv & cal / (mol K) & 0.65 & 0.27 & 0.42 & \textbf{0.15}\tabularnewline
\bottomrule
\end{tabular}
\end{table}

QM9 dataset~\cite{ramakrishnan_quantum_2014} is a widely used comprehensive
dataset that provides geometric, energetic, electronic and thermodynamic
properties of small organic molecules. It consists of 130k molecules
with 12 properties, which are calculated by quantum mechanical simulation
method (DFT). Each molecule consists of Hydrogen (H), Carbon (C),
Oxygen (O), Nitrogen (N), and Fluorine (F). The number of atoms in
each molecule is up to 30. In QM9 dataset, the discrete graph structure
of molecules and atom coordinates are provided, while our model does
not use the discrete graph structure explicitly. For details about
the molecule properties, please refer to \cite{gilmer_neural_2017}.

Table \ref{tab:QM9} shows the experimental results on QM9 dataset.
Our GGRNet consistently outperforms other baselines except U0 and U (atomization energy at 0K and 298.15K), H (enthalpy of atomization), and G (free energy of atomization).
This results show that our model has potential to learn the representations of
small organic molecules in a variety of properties.
The energy of the electron in the highest occupied molecular
orbital (HOMO) and that of the lowest unoccupied molecular orbital
(LUMO) are not improved sufficiently.

\subsection{Ablation Study}

\begin{table}
\caption{\label{tab:ablation}Ablation study on QM9 dataset.}

\centering
\renewcommand{\arraystretch}{1.3}

\begin{tabular}{ccccc}
\toprule 
Model & HOMO & R2 & ZPVE \tabularnewline
\midrule
Full model (GGRNet) & 0.00372 & 1.61 & 0.000861 \tabularnewline
Full model without $\mathbf{x}_{N}$ & 0.00374 & 2.00 & 0.000528 \tabularnewline
Full model without  $\left\{ \mathbf{d}_{v,w}\right\} $ & 0.0144 & 154 & 0.00137 \tabularnewline
Full model without $\left\{ \mathbf{x}_{v}\right\} $ & 0.00441 & 1.82 & 0.000939 \tabularnewline
\bottomrule
\end{tabular}
\end{table}

We performed the ablation study on the subset of QM9 dataset. Table
\ref{tab:ablation} shows the experimental results. The top row is
the original GGRNet and the below is the full model without counting
feature, distance feature, and atom embedding feature, respectively.


The distance feature is indispensable for the accurate prediction
of R2 (an electronic spatial extent). This is also reasonable since
this property reflects the spatial distribution of electrons in the
molecule. Finally, the atom embedding feature is moderately effective
for all properties. Overall, we verified that every additional feature
helps to improve the performance for molecular property prediction.

\section{Conclusions}

In this work, we proposed a GGRNet for accurate and efficient molecular
property prediction. In our model, the parameters for updating hidden
representations are shared across all layers, the input representations
are fed into every layers as skip connections to accelerate the training,
and the hidden representation at each atom is updated by using that
of every other atoms in the graph, which boosts the performance of
the property prediction of molecules. Experiments on the standard
benchmarks of molecular property prediction generated by quantum-chemical
simulations show that our model achieved the state-of-the-art performance
on every datasets. Future work includes applying more expressive functions
for update and readout functions for our model.

\bibliographystyle{plain}
\bibliography{manuscript}

\end{document}